\documentclass{article}
\usepackage{times}

\usepackage{footmisc}

\makeatletter
\renewcommand\maketitle{\par
  \begingroup
    \def\@makefnmark{\rlap{\@textsuperscript{\normalfont\@thefnmark}}}%
    \long\def\@makefntext##1{\parindent 1em\noindent
            \hb@xt@1.8em{%
                \hss\@textsuperscript{\normalfont\@thefnmark}}##1}%
    \if@twocolumn
      \ifnum \col@number=\@ne
        \@maketitle
      \else
        \twocolumn[\@maketitle]%
      \fi
    \else
      \newpage
      \global\@topnum\z@   
      \@maketitle
    \fi
    \thispagestyle{plain}\@thanks
  \endgroup
  \setcounter{footnote}{0}%
  \global\let\thanks\relax
  \global\let\maketitle\relax
  \global\let\@maketitle\relax
  \global\let\@thanks\@empty
  \global\let\@author\@empty
  \global\let\@date\@empty
  \global\let\@title\@empty
  \global\let\title\relax
  \global\let\author\relax
  \global\let\date\relax
  \global\let\and\relax
}
\makeatother

\begin{document}

\date{}
\title{Estimation of fuzzy anomalies in Water Distribution Systems}
\author{J.~Izquierdo$^1$\footnote{jizquier@gmmf.upv.es},
        M.M.~Tung$^2$\footnote{mtung@imm.upv.es},
        R.~Per\'ez$^1$, and
        F.J.~Mart\'{\i}nez$^1$ \\
        $^1$Centro Multidisciplinar de Modelaci\'on de Fluidos \\
        and $^2$Instituto de Matem\'atica Multidisciplinar \\
        Universidad Polit\'ecnica de Valencia, Spain
          }
\maketitle

{\bf Abstract:} State estimation is necessary in  diagnosing  anomalies  in  Water
Demand Systems (WDS). In this paper we present a neural  network  performing
such a task. State estimation is  performed  by  using  optimization,  which
tries to reconcile all the  available  information.  Quantification  of  the
uncertainty of the input data (telemetry measures  and  demand  predictions)
can be achieved by means of robust estate estimation. Using  a  mathematical
model of the network, fuzzy estimated states for  anomalous  states  of  the
network can be obtained. They are used to train a neural network capable  of
assessing WDS anomalies associated with particular sets of measurements. \\

\noindent
{\bf Keywords:} Water Distribution Systems; Neural Nets; Fuzzy Logic; Modelling

\section{Introduction}
\label{sec:1}

Water companies use
telemetry systems for control and operation  purposes.  By  considering  the
data  provided  by  telemetry,  the  engineer  on  duty  makes  operation
decisions trying to  optimize  the  system  utilization.  Nevertheless,  the
system complexity does not permit but  to  take  a  few  real-time  measures,
which only incompletely represent the network state. They give  indication  of  only
certain aspects of the system, leaving out  other  more  specific  or  ``less
relevant'' ones. Thus, suitable techniques that allow for more  accurate  network
health estimation are necessary so  that  anomalies  can  be  detected  more
rapidly, and light anomalies, which develop  progressively  and  insidiously,
can be identified.  This  will  enable  to  control  their  consequences  in
earlier stages, thus avoiding, among other things, losses of water, which can
be  of great importance. 

The state of a WDS is obtained by interrelating different measures within  a
mathematical model of  the  network,
\cite{Mar95}. Different tools to analyze water networks have  been
developed in the last years, SARA~\cite{GMF98}, and EPANET \cite{Ros97},
among others.
But state estimation cannot be accurately performed if there are missing  or
uncertain data. Thus, system operators 
need error limits for the state variables.  Yet,  data  are
abundant since they  are  permanently  received.  Therefore,  operators  cannot
evaluate errors easily or in real time.
It  is  expected  that  suitable   techniques   borrowed   from   Artificial
Intelligence (AI) could encapsulate the necessary knowledge  to  assess  the
network state.

In  this  paper,  we  present  an
approach for the diagnosis and decision making process which is necessary on
a neural network for clustering and  pattern  classification.  
First,  the  mathematical
model,  a  state  estimation  procedure  and  a   mechanism   for   treating
uncertainties, already presented in \cite{Izq04}  and  \cite{Izq05},
are briefly presented.
The state estimator, together with the error limits 
will be used as a  surrogate  of  the  real
WDS to generate data to train and check the neural network (NN).  Then,  the
inherent procedures to neural techniques will  be  described.  Specifically,
the NN architecture, the classification and clustering  mechanisms  of  both,
crisp and fuzzy, patterns and  the  training  technique  will  be  presented.

\section{Mathematical Model and State Estimation}
\label{sec:2}

Analyzing pressurized water  systems  is a 
complex task,  especially  for  big  systems.
But even for moderately sized cities, it involves solving a  big
number of  non-linear  simultaneous  equations.  
The complete set of equations may be written
by using block-matrix notation,
\begin{equation}\label{system}
        \left(
        \begin{array}{cc}
                A_{11}(q) & A_{12} \\
                A_{12}^t  & 0
        \end{array}
        \right)
        \left(
        \begin{array}{c}
                q \\ H
        \end{array}
        \right) =
        \left(
        \begin{array}{c}
                -A_{10} H_f \\ Q
        \end{array}
        \right),
\end{equation}
where $A_{12}$ is the so-called connectivity matrix describing the way demand
nodes are connected through the lines.  Its  size  is  $L\times N_p$,
$N_p$  being  the
number of demand nodes and $L$ the number of lines; $q$ is the vector of the
flow rates through the  lines,
H the vector of unknown heads at demand nodes; $A_{10}$ is an $L\times N_f$
matrix,  $N_f$
being the number of fixed-head nodes with known head $H_f$, and  $Q$  is
the  $N_p$-dimensional vector of demands. Finally, $A_{11}(q)$ is an
$L\times L$  diagonal  matrix.
System (\ref{system}) is a non-linear problem whose  solution  is  the  state  vector
$x=(q,H)^t$ of the system.

The non-linear relations describing the
system balances are complemented by the specific telemetry measurements.
These measurements are integrated into the
model by expanding system (\ref{system}) to a new system, typically
overdetermined:
\begin{equation}\label{system2}
        \left(
        \begin{array}{cc}
                A_{11}(q) & A_{12} \\
                A_{12}^t  & 0      \\
                A_{31}    & A_{32}
        \end{array}
        \right)
        \left(
        \begin{array}{c}
                q \\ H
        \end{array}
        \right) =
        \left(
        \begin{array}{c}
                -A_{10} H_f \\ Q \\M_t
        \end{array}
        \right).
\end{equation}
The components $A_{31}$ and $A_{32}$ in system (\ref{system2}) were
introduced to account for additional telemetry measurements $M_t$ with
uncertainties in the demand predictions.
System (\ref{system2}) is usually solved using least-square methods
for a state estimation by an over-relaxation iterative process applied to
a linearized version of (\ref{system2}):
\begin{equation}\label{system3}
        \left(
        \begin{array}{cc}
                A'_{11}(q^{(k)}) & A_{12} \\
                A_{12}^t  & 0      \\
                A_{31}    & A_{32}
        \end{array}
        \right)
        \left(
        \begin{array}{c}
                \Delta q \\ \Delta H
        \end{array}
        \right) =
        \left(
        \begin{array}{c}
                -A_{10} H_f - A_{11}(q^{(k)})q^{(k)} - A_{12} H^{(k)} \\
                Q - A_{21} q^{(k)} \\
                M_t - A_{31}q^{(k)} - A_{32}H^{(k)}
        \end{array}
        \right),
\end{equation}
where $A'_{11}$ is the Jacobian matrix  corresponding  to  $A_{11}$.

\section{Error Limit Analysis}
\label{sec:3}

Error limit analysis is a process to determine uncertainty  bounds  for  the
state estimation originated by the lack of  precision  of  measurements
and data. 
To put it in a nutshell, the question is what  is  the  reliability  of  the
estimated state $x^*$, if measurement vectors $y$ are not crisp but may vary
in some region, $[y -\delta y, y+\delta y]$?

Different techniques may be  used  to  estimate  this  unknown  but  bounded
error, \cite{Mil96}, \cite{Nor86}, \cite{Kur97}.
We use a variant of the so-called sensitivity matrix analysis, \cite{Bar03},
which uses the state estimator presented above. 

In \cite{Izq05}, it is proved  that  a  component  by
component bound, $e^*$, for $\delta x^*$ can be obtained by means of
\begin{equation}\label{eq7}
        e^* = \big|
        \left( A^{*t}_k W A^*_k \right)^{-1}
        A^{*t}_k W \big|
        \,\big| \delta y \big|,
\end{equation}
where $W$ is a diagonal matrix that weights the equations according to the
nature of the right-hand sides, and the vertical bars indicate absolute
values of all  matrix  and  vector entries. Because of linearity, the bounds
calculated by  (\ref{eq7}) are symmetrical and the error limit may be expressed
as a multidimensional interval (see  cell definition in next section)
$[x^*]$ in the state space
\begin{equation}\label{eq9}
        \left[x^*\right] =
        \big[x^*_{\mathrm{\,inf}}, x^*_{\mathrm{\,sup}}\big] =
        \big[x^*-e^*, x^*+e^*\big].
\end{equation}

\section{The Neural Network}
\label{sec:4}

A neural network for clustering and classification is a mechanism for 
pattern recognition. Here, we use 
multidimensional cells, \cite{Sim92}, \cite{Lik94}. 
Voronoi diagrams are used in Ref.~\cite{Ble97}.

A cell $C$ is a region of the pattern space of $n$-dimensional vectors
obtained as the intersection of $n$ pairs of half-spaces of the form
$m_i\le x_i\le M_i$, for $i = 1,2,\ldots,n$, where $m_i$ and $M_i$ are 
real numbers. Vectors $m = (m_i, i = 1,\ldots,n)$ and $M = (M_i, i = 1, 
\ldots, n)$ are called min and max points of $C$ and completely determine $C$. 
Membership of patterns to a cell is defined from fuzzy grounds. For fuzzy 
patterns, $P=\big[P^{\mathrm{\,inf}},P^{\mathrm{\,sup}}\big]$, like the ones
obtained in (\ref{eq9}), membership values are given by the membership function
\begin{equation}\label{eq10}
        c(P) = \max\limits_{i=1,\ldots,n}
        \Big\{ \max \Big\{
        \varphi_i\left(P_i^{\mathrm{\,sup}}-M_i\right),
        \varphi_i\left(m_i-P_i^{\mathrm{\,inf}}\right)
        \Big\} \Big\},
\end{equation}
where each $\varphi_i(x)$ controls the cell fuzziness.

Values taken by membership function (\ref{eq10}) are used
during the operation phase
to decide the membership degree to the class associated
with a cell exhibited by certain pattern presented to it and, as a consequence,
to recognize the potential anomalous state of the water distribution system
corresponding to the associated label of each class.

Patterns presented to the network during the training phase are ordered 
pairs $(P,l)$, where $l$ is a 
label associated to pattern $P$ describing the type of anomaly it represents. 

The NN implementing the classification process is a three-layer network 
that grows adapting itself to the problem characteristics.
The input layer has $2n$ neurons, two for any of the 
dimensions of the patterns $P=\big[P^{\mathrm{\,inf}},P^{\mathrm{\,sup}}\big]$.
When a new pattern is presented to the network through the input layer, the
components of vectors $P^{\mathrm{\,inf}}$ and $P^{\mathrm{\,sup}}$ are
compared, respectively, with those of the minimum point, $m$, and the
maximum point, $M$, of the $J$ existing cells.
Specifically, numbers in the inner brackets of (\ref{eq10}) are calculated.

This way, each neuron on the hidden layer has two $n$-dimensional vectors 
$\varphi^{\mathrm{\,inf}}$ and $\varphi^{\mathrm{\,sup}}$ as its input,
formed by numbers between $0$ and $1$, ready to be processed, first component
by component with the max operator, and then with the max operator,
but now through all the components. Specifically,
$$
        c(P) = 1 - \max\limits_{i=1,\ldots,n}\Big\{
        \max\Big\{\varphi^{\mathrm{\,sup}}_i,
        \varphi^{\mathrm{\,inf}}_i\Big\}\Big\}
$$
is calculated for each cell. This process 
gives the membership degree of 
$P$ to every one of the cells. Thus, membership functions may be considered as 
the transfer or activation functions for all the $J$ existing hidden neurons. 
And the values of the minimum and maximum points of those existing cells, 
which will be adjusted during the training phase, must be regarded precisely 
as the synaptic weights between the input and the hidden layer.

The values produced by the membership functions of the existing cells 
constitute the outputs of the hidden layer. These values must be operated 
with the weights between the hidden and the output layers. This process 
will produce a class, a diagnosis of the hydraulic system represented by 
pattern $P$. This procedure facilitates the decisions to be made by
the system managers.

\section{Conclusions}
\label{sec:5}

The described neural procedure does not fit into any standard paradigm, since 
it is made of several sub-nets that evolve by accumulating experience as new 
loads (peak, valley, seasonal-dependent, etc.) are observed, which mimics 
human knowledge acquisition. 

From the reduced number of tests performed we conclude that the classification 
ability of the NN is excellent.
Since the response given by the NN is graded, as a consequence of its 
fuzziness, the information it provides is not only qualitative 
(pointing out an anomaly) but also quantitative (weighting the distributed 
importance of the problem). 

The tool presented here, once completed, calibrated and implemented, will 
provide WDS managers with a decision support mechanism allowing early 
identification of anomalies and, as a consequence, better Integrated Water 
Management.

{\bf Acknowledgments.}
This work is been performed under the support of the projects 
Investigaci\'on Interdisciplinar n$^{\mathrm{o}}$ 5706 (UPV), 
\textbf{DPI2004-04430} of the Direcci\'on General de Investigaci\'on del 
Ministerio de Educaci\'on y Ciencia (Spain) and FEDER funds.

\ \vspace{-.7cm}


\end{document}